\documentclass[sigconf]{acmart} 

\AtBeginDocument{%
  \providecommand\BibTeX{{%
    \normalfont B\kern-0.5em{\scshape i\kern-0.25em b}\kern-0.8em\TeX}}}

\copyrightyear{2022}
\acmYear{2022}
\setcopyright{acmcopyright}\acmConference[ICPP '22]{51st International Conference
on Parallel Processing}{August 29-September 1, 2022}{Bordeaux, France}
\acmBooktitle{51st International Conference on Parallel Processing (ICPP '22), August
29-September 1, 2022, Bordeaux, France}
\acmPrice{15.00}
\acmDOI{10.1145/3545008.3545073}
\acmISBN{978-1-4503-9733-9/22/08}

\usepackage{booktabs, multirow} 
\usepackage{amsmath} 
\usepackage{algorithm} 
\usepackage[noend]{algpseudocode} 
\algrenewcommand\algorithmicrequire{\textbf{Input:}}
\algrenewcommand\algorithmicensure{\textbf{Output:}}
\usepackage{subcaption} 
\usepackage{bm} 
\usepackage{tikz}
\newcommand{\cready}{\textcolor{black}}
\begin{document}
\title{FedClassAvg: Local Representation Learning for Personalized Federated Learning on Heterogeneous Neural Networks}

\author{Jaehee Jang}
\email{hukla@snu.ac.kr}
\orcid{0000-0003-0322-5654}
\affiliation{%
  \institution{Department of Electrical and Computer Engineering\\Seoul National University}
  \city{Seoul}
  \country{South Korea}
}

\author{Heonseok Ha}
\email{heonseok.ha@snu.ac.kr}
\orcid{0000-0003-4029-1546}
\affiliation{%
  \institution{Department of Electrical and Computer Engineering\\Seoul National University}
  \city{Seoul}
  \country{South Korea}
}

\author{Dahuin Jung}
\email{annajung0625@snu.ac.kr}
\orcid{0000-0002-1344-1054}
\affiliation{%
  \institution{Department of Electrical and Computer Engineering\\Seoul National University}
  \city{Seoul}
  \country{South Korea}
}

\author{Sungroh Yoon}
\email{sryoon@snu.ac.kr}
\orcid{0000-0002-2367-197X}
\affiliation{
  \institution{Department of Electrical and Computer Engineering\\Interdisciplinary Program in Artificial Intelligence\\Seoul National University}
  \city{Seoul}
  \country{South Korea}
}
\authornote{Corresponding author}

\renewcommand{\shortauthors}{Jang et al.}

\begin{abstract}
Personalized federated learning is aimed at allowing numerous clients to train personalized models while participating in collaborative training in a communication-efficient manner without exchanging private data.
However, many personalized federated learning algorithms assume that clients have the same neural network architecture, and those for heterogeneous models remain understudied.
In this study, we propose a novel personalized federated learning method called federated classifier averaging (FedClassAvg).
Deep neural networks for supervised learning tasks consist of feature extractor and classifier layers.
FedClassAvg aggregates classifier weights as an agreement on decision boundaries on feature spaces so that clients with not independently and identically distributed (non-iid) data can learn about scarce labels.
In addition, local feature representation learning is applied to stabilize the decision boundaries and improve the local feature extraction capabilities for clients. 
While the existing methods require the collection of auxiliary data or model weights to generate a counterpart, FedClassAvg only requires clients to communicate with a couple of fully connected layers, which is highly communication-efficient.
Moreover, FedClassAvg does not require extra optimization problems such as knowledge transfer, which requires intensive computation overhead.
We evaluated FedClassAvg through extensive experiments and demonstrated it outperforms the current state-of-the-art algorithms on heterogeneous personalized federated learning tasks.
\end{abstract}




\maketitle

\section{Introduction}
Federated learning is a privacy-preserving collaborative machine-learning technique.
It enables multiple clients and a global server to train by exchanging knowledge from local training and the data itself.
Because the data distributions of clients are not independent and identically distributed (non-iid), and conventional parallel machine learning algorithms assume iid data distributions of clients, new algorithms are needed.
Beginning with FedAvg~\cite{FedAvg}, many studies~\cite{kairouz2021advances,ji2021emerging} have been proposed to improve the generalization performance of federated learning algorithms.
However, because federated learning concentrates on improving the global model, the client model performance for local data distribution deteriorates.
\begin{figure*}[!ht]
	\centering
	\includegraphics[width=\linewidth]{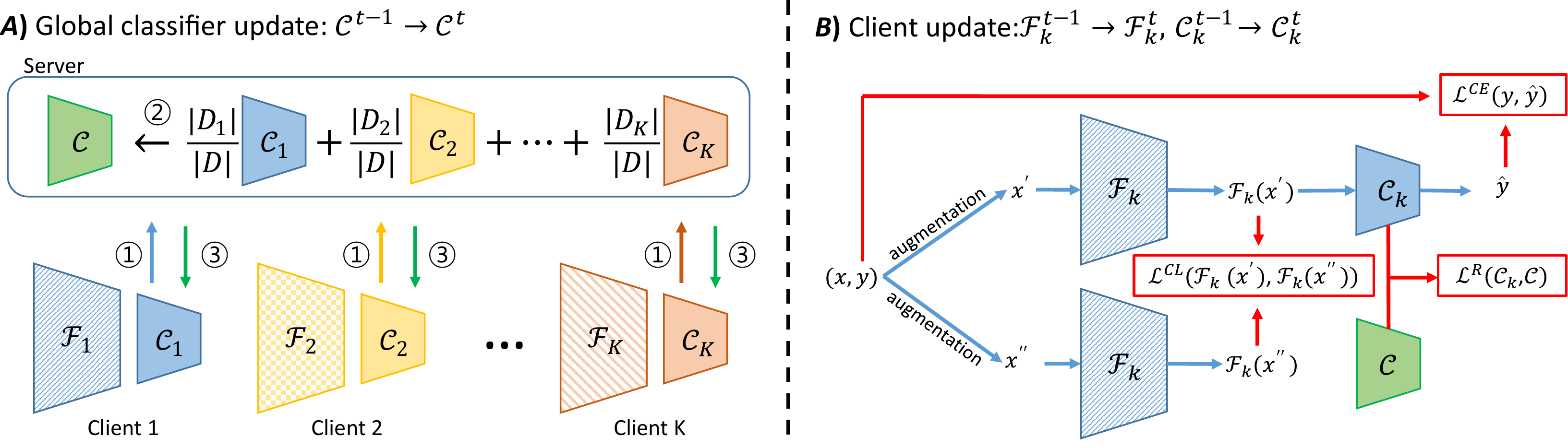}
	\caption{Illustration of FedClassAvg. $\mathcal{F}_\mathrm{*}$ are feature extractors and $\mathcal{C}_\mathrm{*}$ are classifiers. FedClassAvg aggregates client classifiers $\mathcal{C}_k$ and build a global classifier $\mathcal{C}$ as described in A), by the following workflow: 1. clients transmit local classifiers to the server, 2. the transmitted local classifiers are linearly combined as a global classifier, and 3. the global classifier is broadcast to clients. The client models are updated with local feature representation learning ($\mathcal{L}^{CL}$), supervised learning ($\mathcal{L}^{CE}$), and proximal regularization ($\mathcal{L}^{R}$) as described in B).}
	\label{fig:overview}
\end{figure*}

Therefore, the concept of personalized federated learning has been proposed.
It aims for the client to train personalized models collaboratively while maintaining model performance on local data distributions. 
Many personalized federated learning techniques~\cite{kulkarni2020survey,tan2022toward} have significantly contributed to addressing data heterogeneity among clients.
Most personalized federated learning algorithms constrain all clients to use the same model architecture.
However, it is necessary for personalized federated learning to allow clients to choose the different model architectures that are effective for various data distributions of clients.

Several studies~\cite{feddf,fedmd,ktpfl,FedZKT} have resolved model heterogeneity through knowledge transfer.
They have successfully delivered learned knowledge from one client to another by using soft predictions on common public data. 
However, it is a burden for the global server to collect auxiliary data, when it is inaccessible to actual training data distributions that clients possess.
Moreover, it might be infeasible for some tasks in which data privacy is crucial, such as medical or financial data, to require even minimal information on client data distributions.
Furthermore, additional optimization problems for knowledge transfer occur in addition to model training and aggregation, resulting in extra computation overhead.
There has also been a study of heterogeneous personalized federated learning using prototype learning~\cite{FedProto}, but requires models to have the same output shape which highly limts the model choices for clients. 

Therefore, we introduce a novel personalized federated learning framework for heterogeneous models called federated classifier averaging (FedClassAvg).
An overview of the proposed method is presented in Figure~\ref{fig:overview}.
In general, a deep neural network model for a supervised learning task can be divided into a feature extractor and a classifier.
The feature extractors maps input data onto feature spaces, and the classifier determines the decision boundaries between feature space representations of different class labels.
FedClassAvg learns heterogeneous models through classifier weight aggregation.
By unifying the classifier, client models learn the same decision boundary, and different feature extractors learn how the feature space representation should be positioned to fit in the decision boundaries.
Therefore, FedClassAvg enables heterogeneous personalized federated learning without the need for additional data collection and transmission.
Moreover, FedClassAvg does not require computations other than model training or classifier aggregation.
It is communication-efficient because only a couple of fully connected layers are transferred instead of the parameters of the entire model.
In our implementation, the clients in FedClassAvg transfer only 2KB of classifier weights for every communication round.

In addition to classifier aggregation, we applied proximal regularization to reduce the L2 distance between the global and client classifiers.
This reinforces the unified objective of the client models and improves the overall training accuracy.
Moreover, we apply local feature representation learning using a supervised contrastive loss~\cite{contrastive:supcon,contrastive:supcon_lang}.
Feature representation learning through a contrastive loss helps the feature representations of semantically the same data to be closer while different data are farther away.
However, classifier aggregation alone cannot prevent decision boundary drifts caused by client models and data heterogeneity.
Therefore, we use the supervised contrastive loss to distance the feature-space representation of different labels, so that a slight migration of the decision boundary does not flip the labels.

The contributions of this paper and the proposed FedClassAvg are as follows:
\begin{itemize}
    \item We introduced FedClassAvg, a novel framework for personalized federated learning on heterogeneous models, by combining classifier aggregation with local representation learning. 
    It does not require any auxiliary data, or intensive computations other than model training and aggregation.
    \item We evaluated the proposed method using various deep neural network models and datasets. The experimental results suggest that FedClassAvg outperforms state-of-the-art algorithms. 
    \item Through several analyses, we demonstrated that FedClassAvg can convey collaborative knowledge using only classifier aggregations.
\end{itemize}

\section{Related Work}
\subsection{Personalized federated learning for heterogeneous models}
\cready{
After several studies have discovered the possibility of federated learning methods with heterogeneous models~\cite{diao2021heterofl,yu2021fed2,he2020group}, personalized federated learning methods for heterogeneous models using knowledge transfer also have been proposed in the literature~\cite{feddf,fedmd,ktpfl,FedZKT}.
}
Because knowledge distillation or transfer delivers learned information to models with different architectures, they can be successfully applied for model-heterogeneous training.
Among the knowledge-transfer-based personalized federated learning algorithms for heterogeneous models, KT-pFL~\cite{ktpfl} has achieved state-of-the-art performance.
In the KT-pFL, the global server aggregates local soft predictions from clients on broadcast public data.
The server then computes the knowledge transfer coefficient, which determines how much knowledge should be updated from one client to another.

However, the need to collect additional public data remains a significant issue.
In particular, it is necessary to ensure that public and private data have similar distributions.
Hence, it is feasible only in cases where the semantic information of private client data is available.
Knowledge-transfer-based algorithms require local models to train multiple epochs, which can incur a huge computational overhead for clients which are usually assumed as edge devices.
Furthermore, calculating the knowledge coefficient and applying it to local models require additional computation for clients.
Unlike previous studies, FedZKT~\cite{FedZKT} takes a computation-intensive task from clients to global servers by aggregating client model weights, which compromises communication overhead instead.
In summary, current knowledge-transfer-based methods either threaten client-side privacy, burden edge computation resources, or are communication-heavy.
However, FedClassAvg only exchanges a couple of fully connected layers without adopting of auxiliary datasets, which is more communication-efficient.

In addition, FedProto~\cite{FedProto}, using prototype training has also been proposed.
FedProto trains the client feature extractors collaboratively by bringing the prototypes from different clients closer. 
By contrast, our proposed method approximates the feature distribution indirectly through a classifier.

\subsection{Feature representation learning}
Self-supervised learning (SSL)~\cite{contrastive:simclr,contrastive:moco,contrastive:byol,contrastive:simsiam} has shown rapid growth in image domains, closing the gap with supervised learning. 
SSL has demonstrated that it is possible to learn useful representations solely using training signals from auxiliary tasks that do not rely on labels. 
SSL applies diverse transformations to the data and utilizes them as supervision to learn representations that are invariant to distortions of the data. 
This promotes learning embeddings that discard nuisance variables, resulting in improved generalization. 
This is commonly achieved by a contrastive loss, minimizing the distance between representations from differently distorted versions while maximizing that of the rest of the instances. 

Recently, several works have extended the contrastive loss to fully supervised scenarios by leveraging both ground-truth labels and self-supervision information~\cite{contrastive:supcon,contrastive:supcon_lang}. 
In contrast to the original self-supervised contrastive loss, supervised contrastive loss pulls all differently augmented instances from the same class and pushes the rest of them in the batch. 
This approach shows consistent superiority over cross-entropy loss on several benchmarks. 
Our work also utilizes the supervised contrastive loss to control the distances between the same class and the different classes, resulting in improved performance.

\section{Methodology}
\subsection{Problem formulation}
In federated learning, each of the $K$ clients learns local model weight $w_{k}$ with non-iid local data $D_{k}$ to improve the performance of the global model.
Hence, the goal of federated learning can be formulated as follows:
\begin{equation}
	\min_{w}\mathcal{L}(f(w;\textbf{x}), \textbf{y}):=\sum_{k=1}^{K}\frac{|D_{k}|}{|D|}\mathcal{L}(f(w_{k};\textbf{x}), \textbf{y}),
\end{equation}
where $L$ is loss function, $f$ is the model architecture, $w$ is the global model weight parameter and $w_k$ is the model weight parameter of client $k$.
$D=(\textbf{x}$, $\textbf{y})$ denotes the oracle dataset, i.e., when the client data are aggregated, where $\textbf{x}$ is input data and $\textbf{y}$ is the corresponding label.

However, personalized federated learning for heterogeneous model aims to minimize clients' average loss at local data distributions.
For client loss $\mathcal{L}_{k}$, data $(\textbf{x}_{k}, \textbf{y}_{k})$ and model architecture $f_{k}$, the objective of personalized federated learning can be formulated as follows:
\begin{equation}
	\min_{w_{1}, \dotsc, w_{K}}\sum_{k=1}^{K}\frac{|D_{k}|}{|D|}\mathcal{L}_{k}(f_{k}(w_{k};\textbf{x}_{k}), \textbf{y}_{k}).
\end{equation}
\subsection{FedClassAvg algorithm}
In this section, the FedClassAvg algorithm is introduced.
Algorithm~\ref{alg:fedclassavg} describes the workflow of the proposed algorithm.
\begin{algorithm}[!ht]
	\caption{FedClassAvg Algorithm}\label{alg:fedclassavg}
	\begin{algorithmic}[1]
		\Require $D_{1}, D_{2}, \dotsc D_{K} \in D, T, E$ 
        \Comment $T$: Maximum communication rounds, $E$: local training epochs
		\Ensure $w_{1}^{t}, \dotsc, w_{K}^{t}$, where $w_k=w_{\mathcal{F}_k}+w_{\mathcal{C}_k}$ 
		\State Initialize $w_{1}^0, \dotsc, w_{k}^0$
		\Procedure {GlobalClassifierUpdate}{}
		\For {Each communication round $t=1, \dotsc, T$}
            \State Sample clients to participate in round $t$
            \State $w_{\mathcal{C}_{k}}^{t} \leftarrow ClientUpdate(w_{\mathcal{C}}^{t})$
            \State $w_{\mathcal{C}}^{t+1} \leftarrow \sum_{k=1}^{K} \frac{|D_{k}|}{|D|} w_{\mathcal{C}_{k}}^{t}$ \Comment Classifier averaging
		\EndFor
		\EndProcedure
		\Procedure {ClientUpdate}{$w_{\mathcal{C}}^{t}$}
            \For {Each local epoch $i$ from 1 to $E$}
            \For {Mini-batch $b \in D_{k}$}
            \State Update model parameters via (\ref{eq:local})
            \EndFor
            \EndFor \\
		\Return Local classifier parameters $w_{\mathcal{C}_{k}}^{t}$
		\EndProcedure
	\end{algorithmic}
\end{algorithm}

First, client models are defined to have different architectures $f_{k}=\mathcal{C}_{k}\circ \mathcal{F}_{k}$ except the classifier part $\mathcal{C}_{k}$.
At the beginning of each communication round, clients are sampled to participate in federated learning.
The number of clients are determined by a predefined client sampling rate, and it remains the same at every communication round. 
Thereafter, the server distributes the global classifier $\mathcal{C}$ for participating clients in the current round.
Subsequently, the clients replace the local classifier with the received $\mathcal{C}$ and train the local model.
After the local weight updates, the global server aggregates the classifier weights $\mathcal{C}_{k}$ to update the global classifier.
\subsubsection{Classifier averaging}
Classifier averaging aggregates classifier weights from local client models.
This can be formulated as follows:
\begin{equation}
    w_{\mathcal{C}}^{t+1} \leftarrow \sum_{k=1}^{K} \frac{|D_{k}|}{|D|} w_{\mathcal{C}_{k}}^{t}.
\end{equation}
Having a unified classifier overcomes data and model heterogeneity across different clients because it provides a common objective.
Taking convolutional neural networks as an example, the convolutional layers are used to extract feature representations of images, and the rear fully connected layers decide which class the feature representation belongs to.
Most deep neural networks can be divided into feature extractors and classifiers; for any model structure, fully connected layers as output layers.
Therefore, in FedClassAvg, we slightly modify the fully connected layers to be shared across the different models.

While implementing FedClassAvg, we adapted different convolutional neural networks such as ResNet~\cite{ResNet}, ShuffleNetV2~\cite{ShuffleNetV2}, GoogLeNet~\cite{GoogLeNet}, and AlexNet~\cite{AlexNet} to share weights by defining the feature extractor $\mathcal{F}_{k}$ as convolutional layers followed by a single fully connected layer and another fully connected layer $\mathcal{C}_{k}$ as a classifier.
Therefore, the client models $f_{k}=\mathcal{C}_{k}\circ\mathcal{F}_{k}$ are constructed.
To create a global classifier based on classifiers with a unified structure, we linearly combine the classifier weights to achieve global classifier weights.

\subsubsection{Local model update with feature representation learning}
After classifier averaging, client models receive global classifier weights from the global server to initialize their classifier weights.
Then, clients train their models using their local training data.
To alleviate the conflict between different classifiers through classifier averaging, we facilitate supervised feature representation learning using a supervised contrastive loss~\cite{contrastive:supcon}.
The supervised contrastive loss directs a model to learn features with the same labels to be close, whereas features with different labels are far. 
This stabilizes the decision boundaries determined by the classifiers.
If a local decision boundary moves slightly through local heterogeneity, the distance between different labels would prevent the label from flipping.

Local feature representation learning is shown in Figure~\ref{fig:overview} (b).
Input data $x$ is perturbed as $x^\prime$ and $x^{\prime\prime}$.
Then the feature extractor $\mathcal{F}$ extracts features from each perturbed input.
Therefore, the final objective of local client update can be formulated as follows:
\begin{equation}\label{eq:local}
	\mathcal{L}_{k}(f_k(w_k;x),y) = \mathcal{L}^{CL}(\mathcal{F}_{k}(x^{\prime}), \mathcal{F}_{k}(x^{\prime\prime})) + \mathcal{L}^{CE}(y, \hat{y}) + \rho \mathcal{L}^{R}(\mathcal{C}, \mathcal{C}_{k}),
\end{equation}
where $\mathcal{L}^{CL}$ is the supervised contrastive loss~\cite{contrastive:supcon}, $x^{\prime}$ and $x^{\prime\prime}$ are augmented inputs of $x$, $\mathcal{L}^{CE}$ is cross entropy loss, $\hat{y}$ is predicted label of input $x^{\prime}$, $\rho$ is regularization ratio and $\mathcal{L}^{R}$ is the L2 distance between client classifier $\mathcal{C}_{k}$ and global classifier $\mathcal{C}$ weights.
Where $w_{\mathcal{C}}$ is the global classifier weight, $\mathcal{L}^{R}$ can be formulated as follows:
\begin{equation}
    \mathcal{L}^R(\mathcal{C}_k, \mathcal{C}) = \lVert w_{\mathcal{C}_k} - w_{\mathcal{C}} \rVert_{2}.
\end{equation}
Using $\mathcal{L}^{CL}$, the extracted features $\mathcal{F}(x^\prime)$ and $\mathcal{F}(x^{\prime\prime})$ are optimized to being closer while growing apart from features of the other classes.
In addition, since local training after global classifier update might cause too much drift from the agreed classifier weights, we add a proximal regularization term similar to FedProx~\cite{FedProx} but only applied for classifiers.

\section{Experimental Results}
\subsection{Experimental setup}
We evaluate the proposed algorithm by measuring average test accuracy on three benchmark datasets: CIFAR-10~\cite{CIFAR10}, Fashion-MNIST~\cite{Fashion-MNIST} and EMNIST~\cite{EMNIST} Letters dataset.
CIFAR-10 is a dataset consisting of 60,000 color images: 50,000 for training images and 10,000 for testing images.
Each data instance is a 32$\times$32 color image of 10 disjoint classes: airplanes, automobiles, birds, cats, deers, dogs, frogs, horses, ships, and trucks.
Fashion-MNIST comprises 28$\times$28 grayscale images in 10 classes with 60,000 examples of a training set and a test set of 10,000 examples.
EMNIST comprises a set of handwritten character images that provide different splits.
In this study, we used the EMNIST Letters split, which has of 145,600 characters with 26 balanced classes consisting of 124,800 training images and 20,800 test images.

\begin{figure}[!htp]
	\begin{subfigure}{.49\linewidth}
		\centering
		\includegraphics[width=\linewidth]{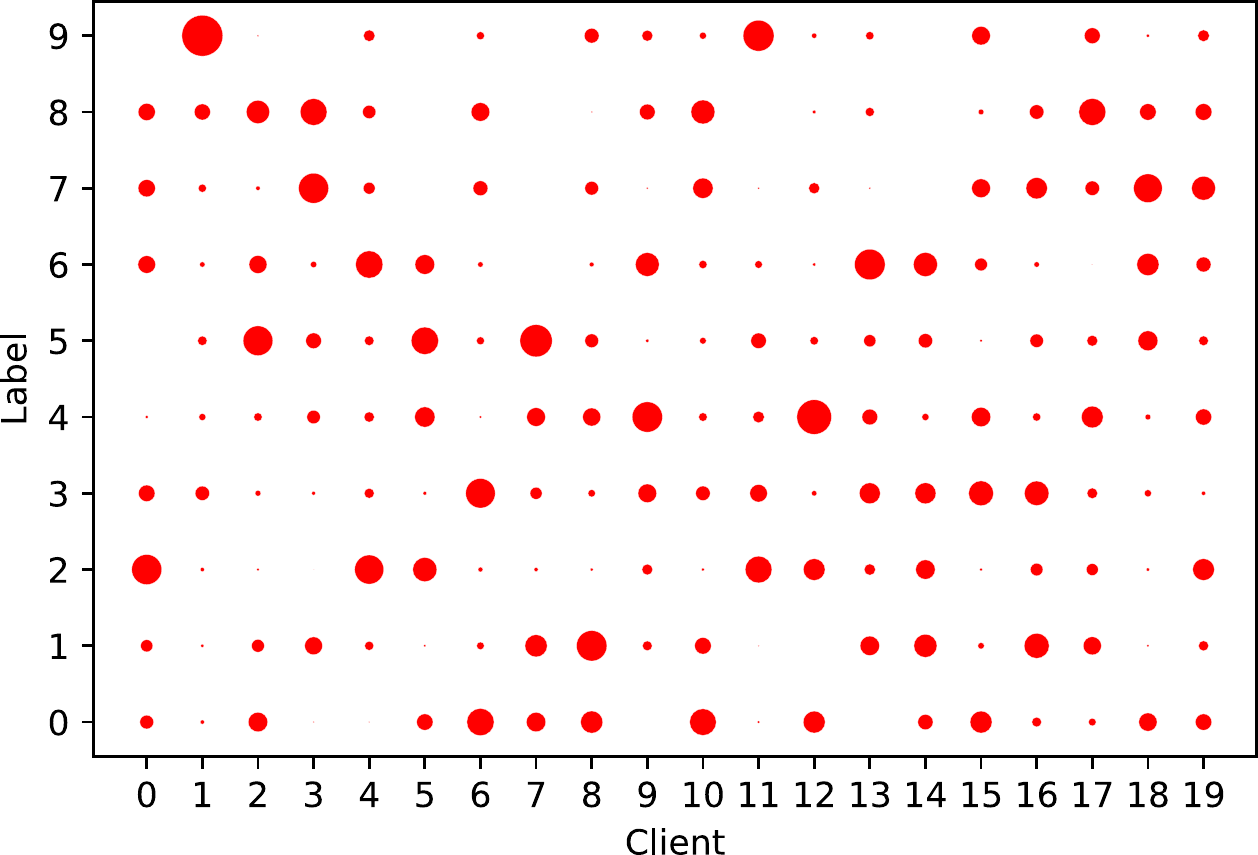}
		\caption{Dir(0.5)}
		\label{fig:cifar10_labeldir}
	\end{subfigure}
	\begin{subfigure}{.49\linewidth}
		\centering
		\includegraphics[width=\linewidth]{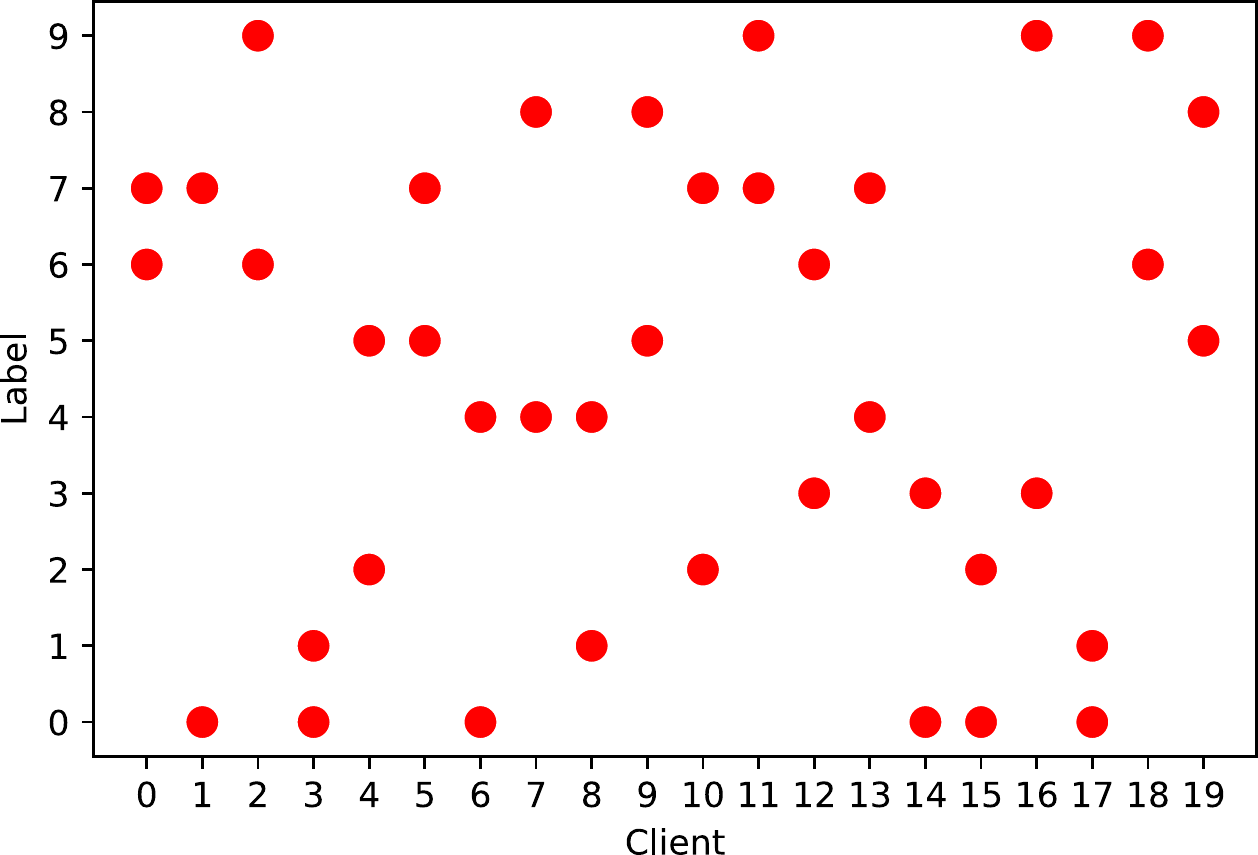}
		\caption{Skewed}
		\label{fig:cifar10_twoclass}
	\end{subfigure}
	\caption{Non-iid label distribution across clients in the CIFAR-10 dataset. Fashion-MNIST is also similarly distributed.}
	\label{fig:cifar10_noniid}
\end{figure}

\begin{figure}[!htp]
\begin{subfigure}{.49\linewidth}
    \centering
    \includegraphics[width=\linewidth]{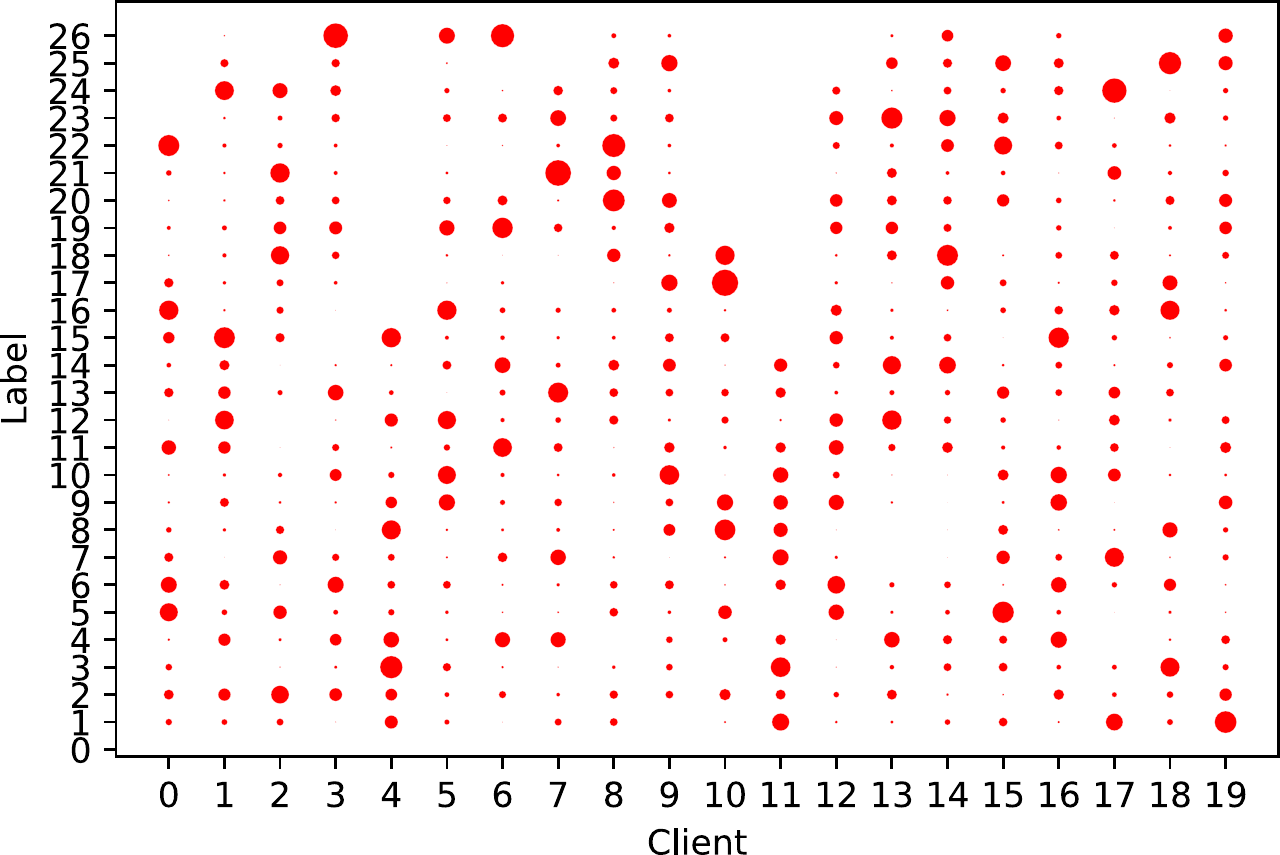}
    \caption{Dir(0.5)}
    \label{fig:emnist_labeldir}
\end{subfigure}
\begin{subfigure}{.49\linewidth}
    \centering
    \includegraphics[width=\linewidth]{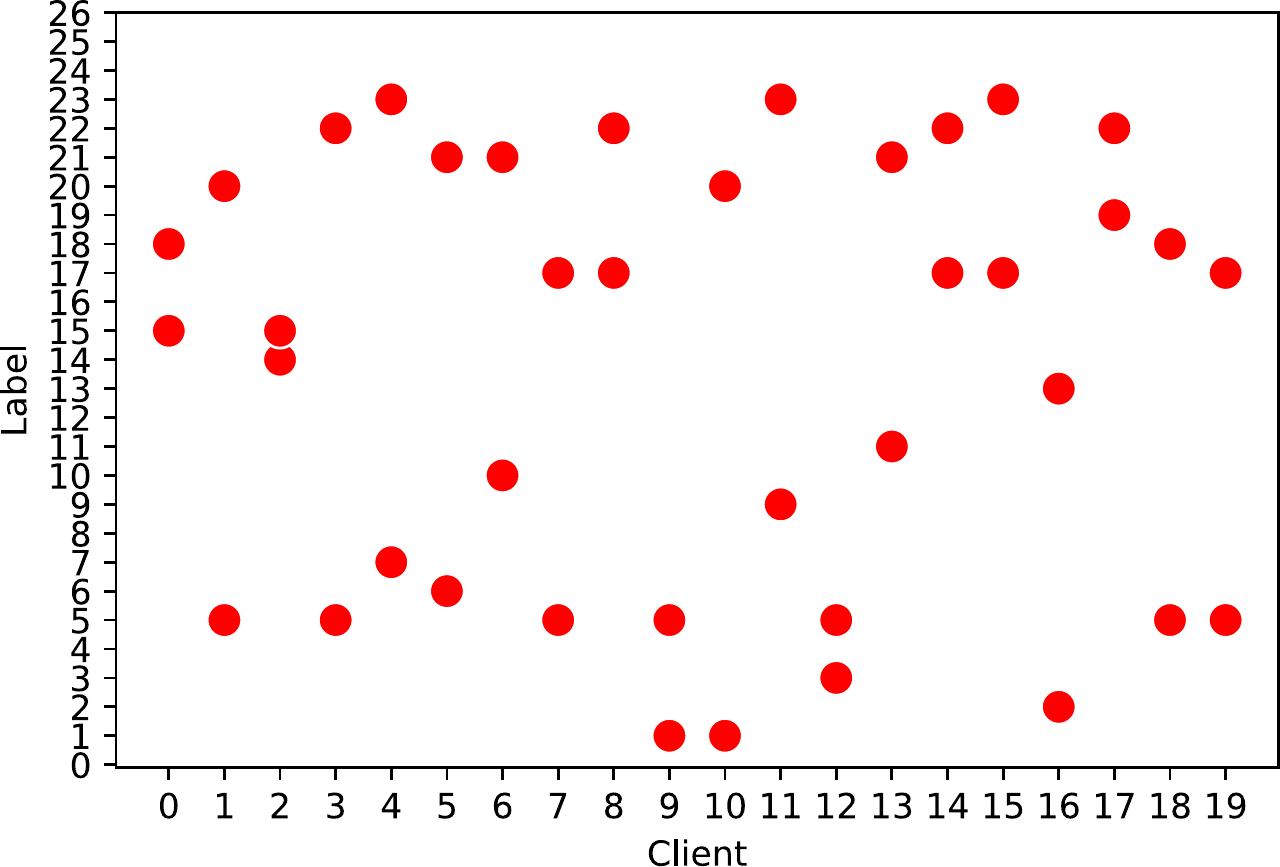}
    \caption{Skewed}
    \label{fig:emnist_twoclass}
\end{subfigure}
\caption{Non-iid label distribution across clients in the EMNIST dataset.}
	\label{fig:emnist_noniid}
\end{figure}

To simulate the data heterogeneity of federated learning environments, we sampled client datasets in two ways: 1) sampling classes using Dirichlet distribution, and 2) skewed class distribution where each client holds datasets with only two sampled classes.
The data sizes of all clients were equally distributed; the resulting data distribution is shown in Figures~\ref{fig:cifar10_noniid} and \ref{fig:emnist_noniid}.
%

For the evaluation models, we used four convolutional neural network architectures: ResNet-18~\cite{ResNet}, ShuffleNetV2~\cite{ShuffleNetV2}, GoogLeNet~\cite{GoogLeNet}, and AlexNet~\cite{AlexNet}.
The models were equally distributed among the clients.
For each model, we used backbone convolutional layers and one following fully connected layer as the feature extractor.
To enable classifier weight averaging, the feature dimensions of all the feature extractors were set to 512.
We used one fully connected layer as the classifier of each model, which had an input dimension of 512 and an output of the number of classes (that is, 10 for CIFAR-10, Fashion-MNIST and 26 for EMNIST).
For model implementations, we used torchvision package from PyTorch\footnote{https://pytorch.org/vision/stable/models.html} without pre-trained weights.
However, we used a custom implementation for AlexNet because the torchvision implementation is only valid for ImageNet datasets.
\cready{
We used ten NVIDIA GeForce 1080Ti GPUs and five 2080Ti GPUs for training. 
20 clients were simulated using a node for every two or four clients on ten 1080Ti GPUs or five 2080Ti GPUs communicating with MPI.
}

The hyperparameters used for local client updates are summarized in Table~\ref{tab:hyperparam}.
\cready{
They were selected through Bayesian hyperparameter optimization. 
The influence of each hyperparameter is as follows: if the learning rate or regularization term $\rho$ becomes large or small, it can cause underfitting/overfitting of the local model, resulting in the model performance degradation. 
In the case of batch size, if it is not too large, the influence of batch size is insignificant. 
The smaller the local epoch, the more frequent communication is required. 
Further, it can prevent the client classifiers from moving too far away from each other so that the global classifier can converge easily.
}
%
\begin{table}[!htp]\centering
	\caption{Hyperparameters used for local client update.}\label{tab:hyperparam}
	\begin{tabular}{lcccc}\toprule
		Dataset       & Learning rate & Batch size & $\rho$ & \# epochs \\\midrule
		CIFAR-10      & 0.0001        & 64         & 0.1    & 1     \\
		Fashion-MNIST & 0.0006        & 64         & 0.4662 & 1     \\
		EMNIST        & 0.0005        & 64         & 0.1    & 1     \\
		\bottomrule
	\end{tabular}
\end{table}
%

We compared our proposed method with KT-pFL~\cite{ktpfl}.
We used the public implementation of KT-pFL\footnote{https://github.com/cugzj/KT-pFL} and FedProto\footnote{https://github.com/yuetan031/fedproto} for a fair evaluation but tested it with our data distribution and models.
The hyperparameters required for the KT-pFL were used as specified by the authors.
\subsection{Heterogeneous federated learning}
\begin{table*}[!htp]\centering
	\caption{Average test accuracies and standard deviations on different datasets on 20 clients with heterogeneous models, i.e., ResNet-18, ShuffleNetV2, GoogLeNet, AlexNet. Models were equally distributed. We assumed two non-iid data distributions: \textit{Dir(0.5)} for Dirichlet label distribution and \textit{Skewed} for data distribution which clients only contain two classes. Bold font indicates best result obtained.}
	\label{tab:hetero_testacc}
	\begin{tabular}{ccccccccc}
        \toprule
		\textbf{Method}          & \multicolumn{2}{c}{\textbf{CIFAR-10}}       & \multicolumn{2}{c}{\textbf{Fashion-MNIST}}  & \multicolumn{2}{c}{\textbf{EMNIST}}                                                                                                                        \\
        \cmidrule(lr){2-3} \cmidrule(lr){4-5} \cmidrule(lr){6-7}
		                         & Dir(0.5)                           & Skewed                             & Dir(0.5)                           & Skewed                             & Dir(0.5)                           & Skewed                             \\\midrule
		Baseline                 & \multirow{2}{*}{$0.6894\pm0.0665$} & \multirow{2}{*}{$0.8871\pm0.0417$} & \multirow{2}{*}{$0.8840\pm0.0698$} & \multirow{2}{*}{$0.9430\pm0.0288$} & \multirow{2}{*}{$0.9149\pm0.0378$} & \multirow{2}{*}{$0.9671\pm0.1073$} \\
		(local training)         &                                    &                                    &                                    &                                    &                                    &                                    \\
		FedProto~\cite{FedProto} & $0.4742\pm0.0323$                  & $0.8359\pm0.0181$                  & $0..6042\pm0.0500$                 & $0.6364\pm0.0423$                  & $0.2249\pm0.0210$                  & $0.2183\pm0.0661$                  \\
		KT-pFL~\cite{ktpfl}      & $0.6228\pm0.0783$                  & $0.8721\pm0.0717$                  & $0.9039\pm0.0435$                  & $0.9737\pm0.0434$                  & $0.9055\pm0.0500$                  & $0.9921\pm0.0076$                  \\
		\textbf{Proposed}        & \bm{$0.7670\pm0.0532$}         & \bm{$0.9202\pm0.0562$}         & \bm{$0.9303\pm0.0308$}         & \bm{$0.9800\pm0.0281$}         & \bm{$0.9305\pm0.0178$}         & \bm{$0.9957\pm0.0040$}         \\
		\bottomrule
	\end{tabular}
\end{table*}
\begin{figure*}[!htp]
	\begin{subfigure}{.33\textwidth}
		\includegraphics[width=\linewidth]{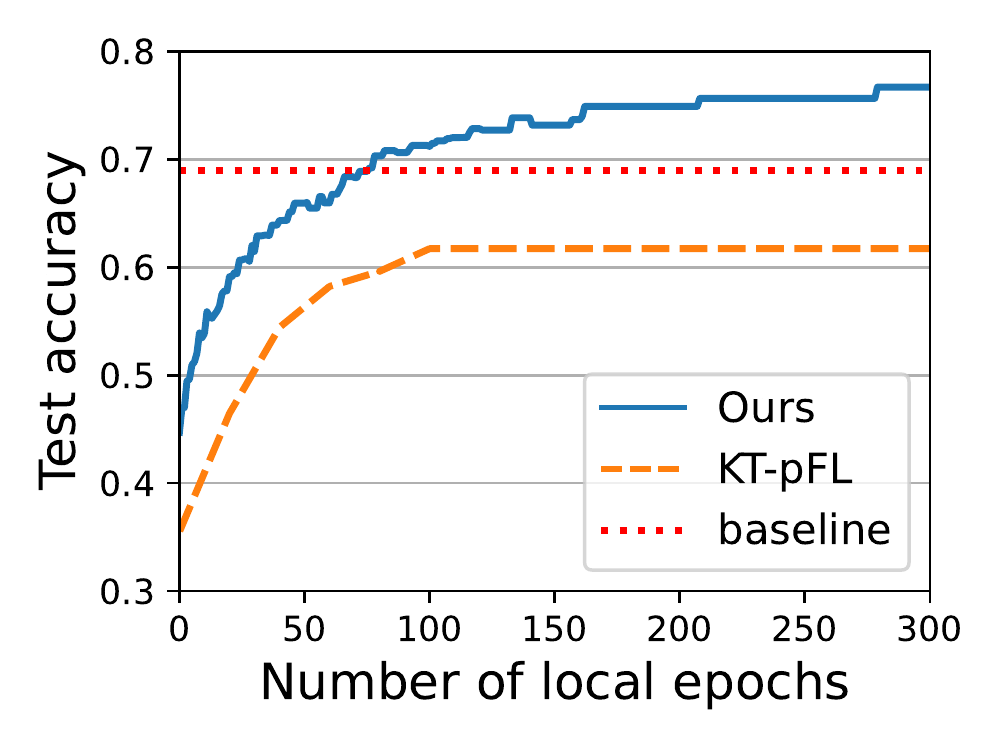}
		\caption{CIFAR-10}
	\end{subfigure}
	\begin{subfigure}{.33\textwidth}
		\includegraphics[width=\linewidth]{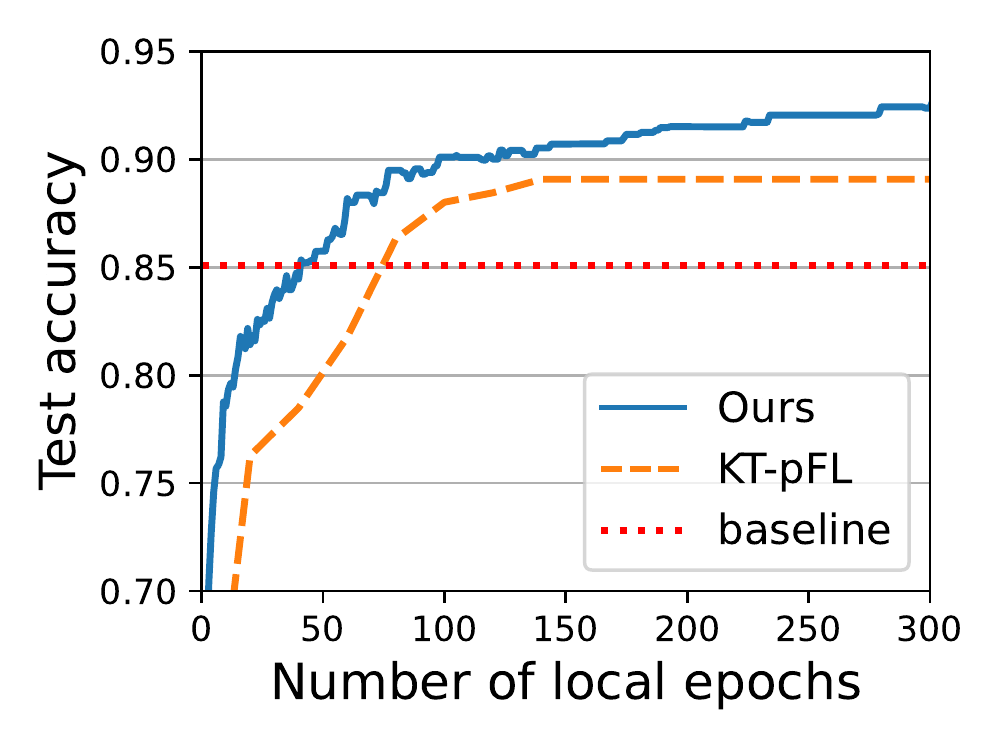}
		\caption{Fashion-MNIST}
	\end{subfigure}
	\begin{subfigure}{.33\textwidth}
		\includegraphics[width=\linewidth]{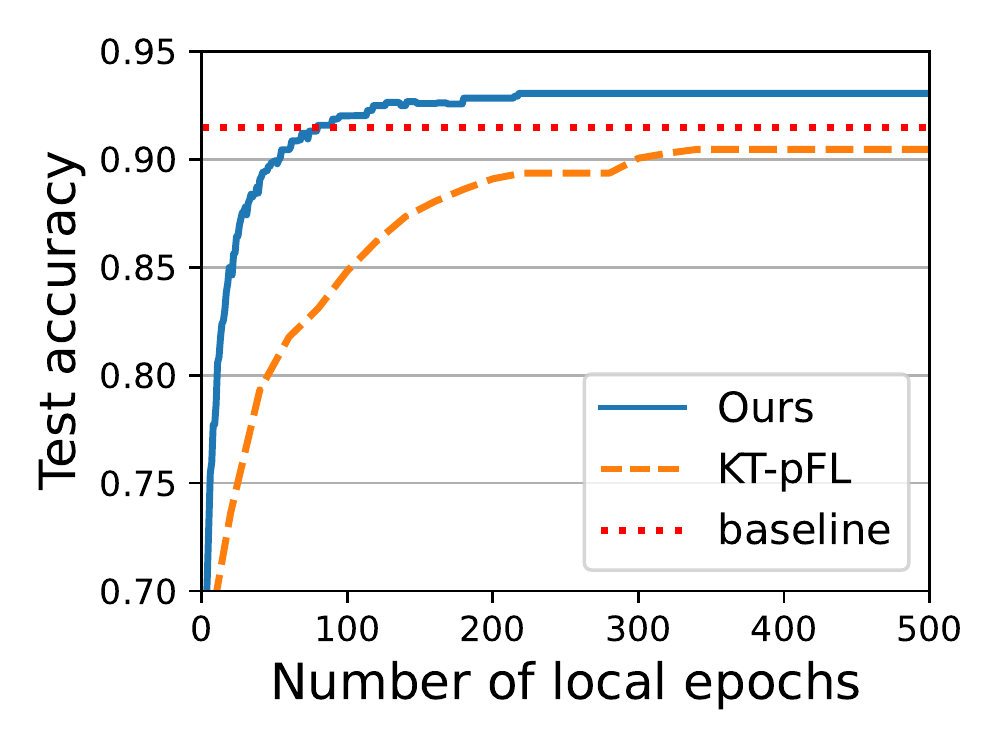}
		\caption{EMNIST}
	\end{subfigure}
	\caption{Learning curves of heterogeneous model training when there are 20 clients with non-iid labels distributed by Dir(0.5).}
	\label{fig:hetero_labeldir}
\end{figure*}
%
\begin{figure*}[!htp]
	\begin{subfigure}{.33\textwidth}
		\includegraphics[width=\linewidth]{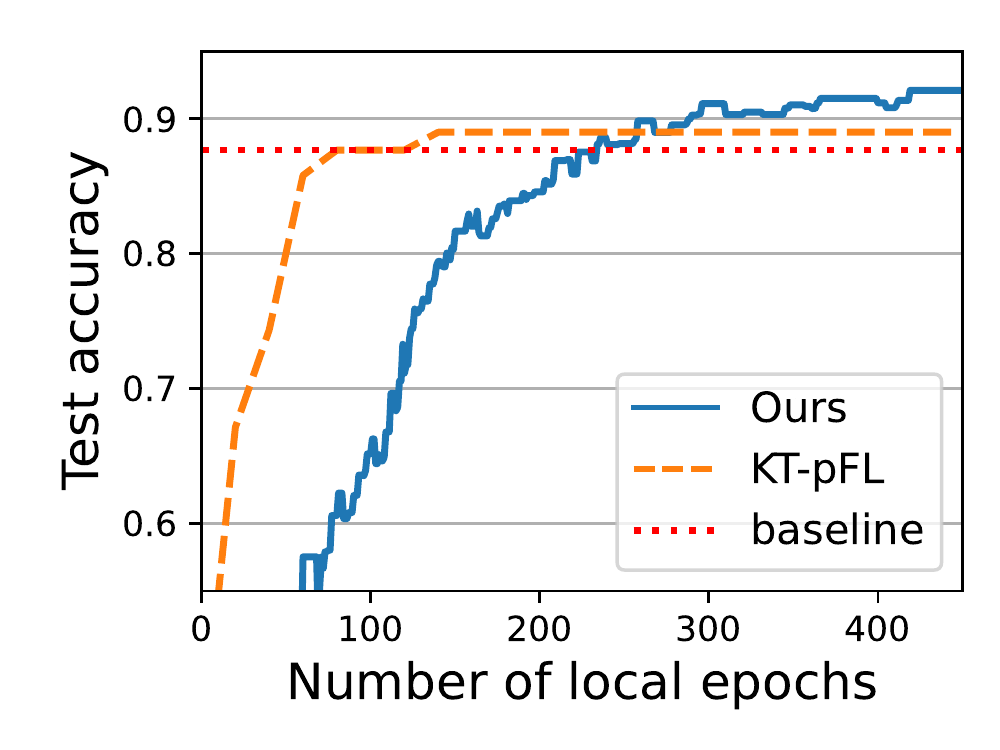}
		\caption{CIFAR-10}
        \label{fig:hetero_twoclass_cifar10}
	\end{subfigure}
	\begin{subfigure}{.33\textwidth}
		\includegraphics[width=\linewidth]{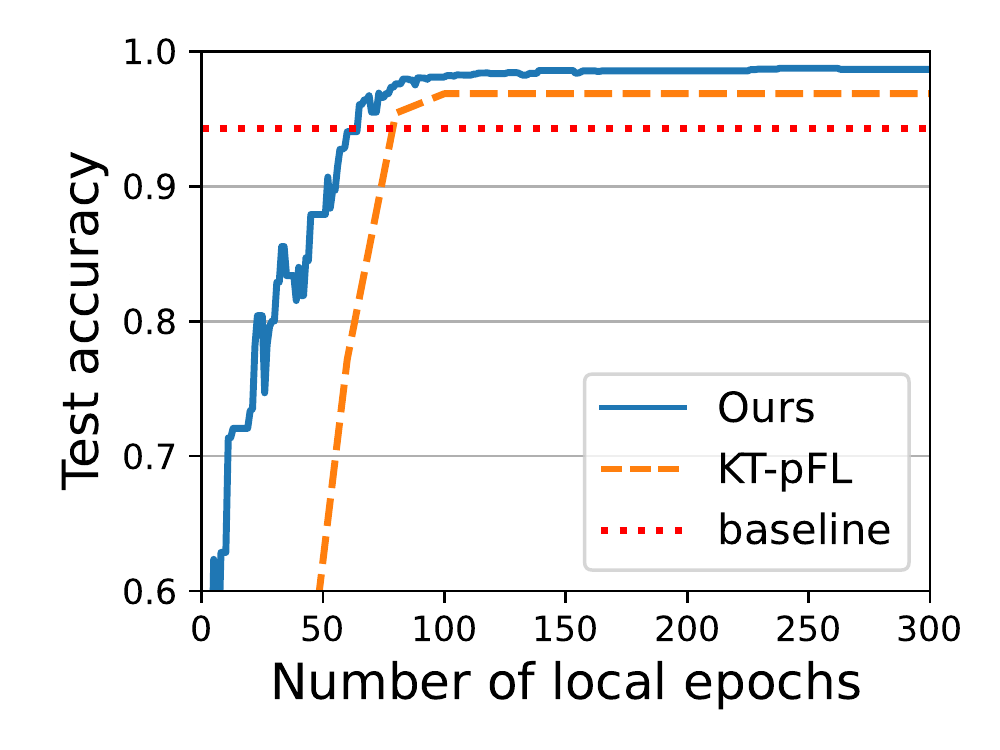}
		\caption{Fashion-MNIST}
	\end{subfigure}
	\begin{subfigure}{.33\textwidth}
		\includegraphics[width=\linewidth]{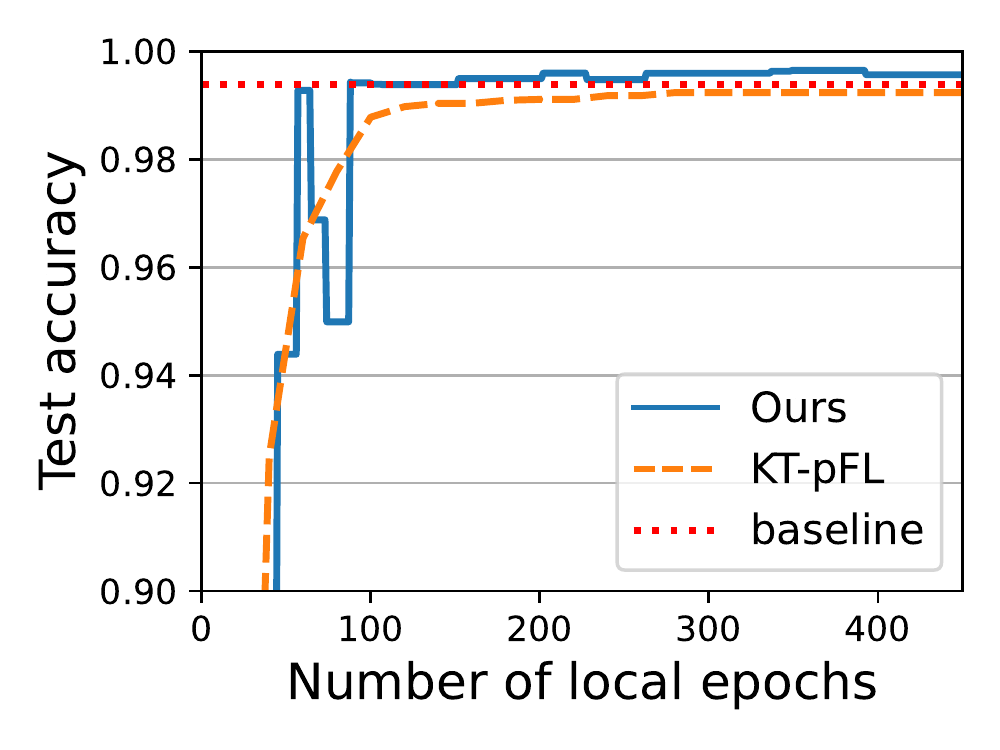}
		\caption{EMNIST}
	\end{subfigure}
	\caption{Learning curves of heterogeneous model training when there are 20 clients containing only two labels.}
	\label{fig:hetero_twoclass}
\end{figure*}
%
\begin{table*}[!ht]\centering
	\caption{Average test accuracies and standard deviations on different datasets with homogeneous models. FedAvg and KT-pFL measured on the same convolutional architecture as FedAvg paper, and ours measured on ResNet-18. For large-scale FL setting with 100 clients, sampling rate was set as $0.1$. +weight indicates when the algorithms shared other weight parameters as well.}\label{tab:homo_testacc}
	\begin{tabular}{lcccccccc}\toprule
		\multicolumn{2}{c}{\textbf{Method}} & \multicolumn{2}{c}{\textbf{CIFAR-10}} & \multicolumn{2}{c}{\textbf{Fashion-MNIST}} & \multicolumn{2}{c}{\textbf{EMNIST}}                                                                         \\
        \cmidrule(lr){3-4} \cmidrule(lr){5-6} \cmidrule(lr){7-8}
		                                    &                                       & 20 clients                                 & 100 clients                         & 20 clients      & 100 clients     & 20 clients      & 100 clients     \\\midrule
\textbf{FedAvg~\cite{FedAvg}} &\textbf{} &$0.7729\pm0.0501$ &$0.6336\pm0.1788$ &$0.8988\pm0.0359$ &$0.7471\pm0.1554$ &$0.9343\pm0.0247$ &$0.8662\pm0.0445$ \\
\textbf{FedProx~\cite{FedProx}} &\textbf{} &$0.8123\pm0.0363$ &$0.6505\pm0.1809$ &$0.9025\pm0.0218$ &$0.7477\pm0.1566$ &$0.9462\pm0.1566$ &$0.8677\pm0.0923$ \\
\textbf{KT-pFL~\cite{ktpfl}} &\textbf{} &$0.5433\pm0.0988$ &$0.4777\pm0.1978$ &$0.8954\pm0.045$ &$0.6114\pm0.2721$ &$0.8505\pm0.0443$ &$0.6589\pm0.3258$ \\
\textbf{} &\textbf{+weight} &$0.6809\pm0.0626$ &$0.5624\pm0.1909$ &$0.9113\pm0.041$ &$0.8647\pm0.066$ &$0.6774\pm0.2311$ &$0.8441\pm0.0316$ \\
\textbf{Proposed} &\textbf{} &$0.7653\pm0.0454$ &$0.5096\pm0.1661$ &$0.9294\pm0.0304$ &$0.6712\pm0.1605$ &$0.9361\pm0.0168$ &$0.7097\pm0.1022$ \\
\textbf{} &\textbf{+weight} &\bm{$0.8546\pm0.0406$} &\bm{$0.7817\pm0.0649$} &\bm{$0.9361\pm0.0158$} &\bm{$0.9057\pm0.0472$} &\bm{$0.9464\pm0.0136$} &\bm{$0.9166\pm0.0275$} \\
		\bottomrule
	\end{tabular}
\end{table*}
%
\begin{figure*}[!ht]
	\begin{subfigure}{.33\textwidth}
		\includegraphics[width=\linewidth]{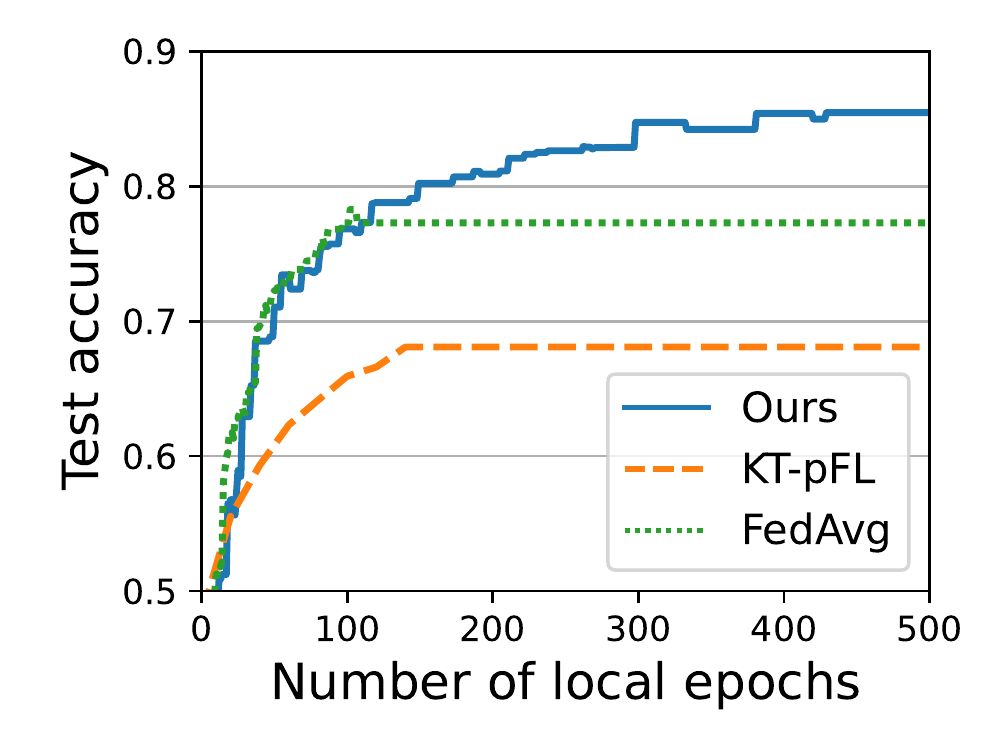}
		\caption{CIFAR-10}
	\end{subfigure}
	\begin{subfigure}{.33\textwidth}
		\includegraphics[width=\linewidth]{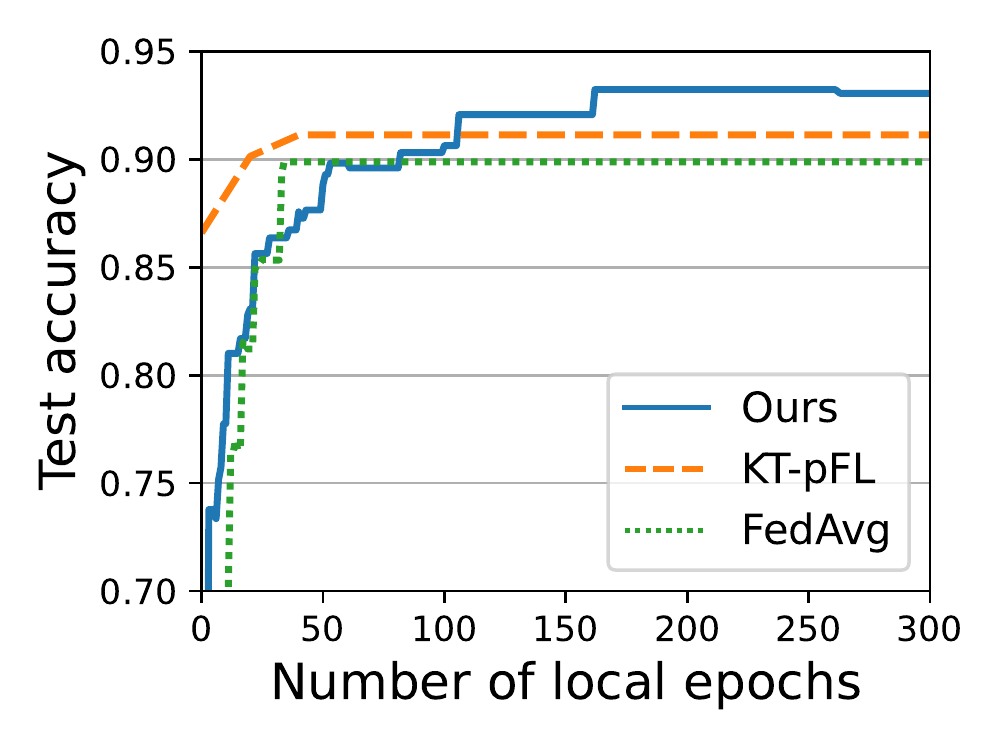}
		\caption{Fashion-MNIST}
	\end{subfigure}
	\begin{subfigure}{.33\textwidth}
		\includegraphics[width=\linewidth]{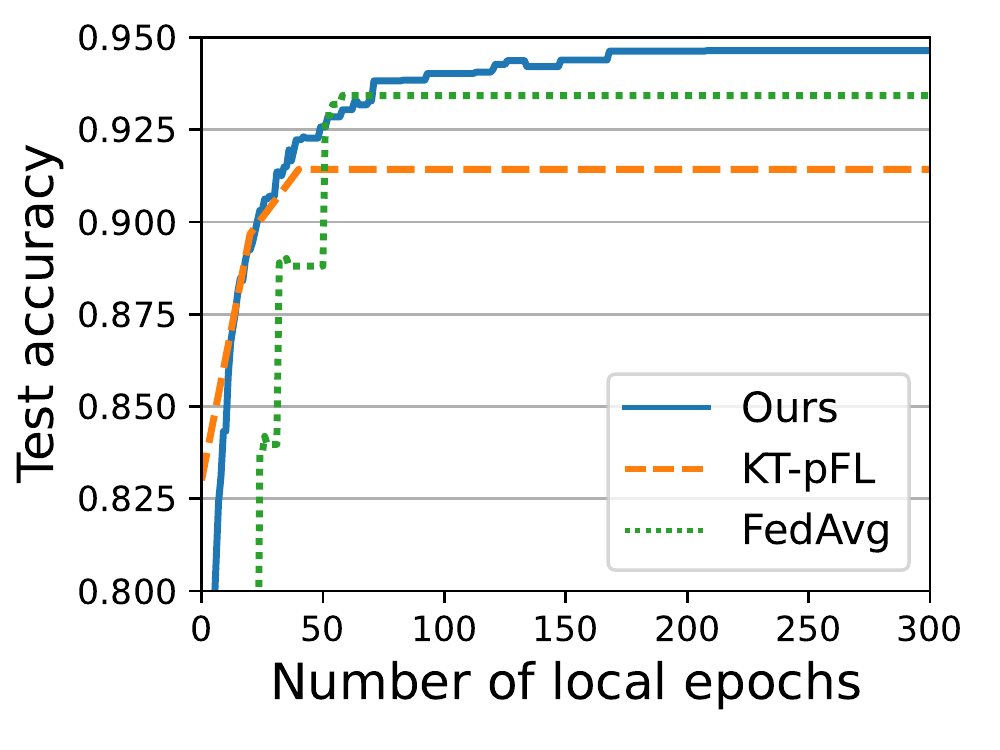}
		\caption{EMNIST}
	\end{subfigure}
	\caption{Learning curves of homogeneous models when clients have non-iid distribution by Dirichlet distribution (Dir(0.5)).}
	\label{fig:homo_20}
\end{figure*}
%
\begin{figure*}[!htp]
	\begin{subfigure}{.33\textwidth}
		\includegraphics[width=\linewidth]{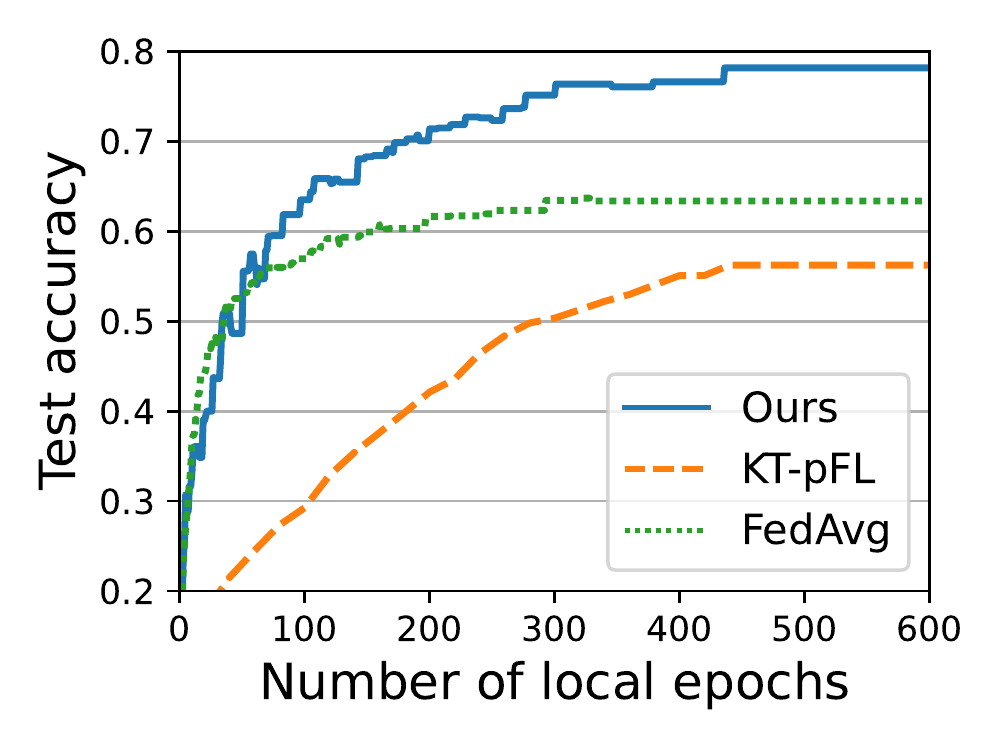}
		\caption{CIFAR-10}
	\end{subfigure}
	\begin{subfigure}{.33\textwidth}
		\includegraphics[width=\linewidth]{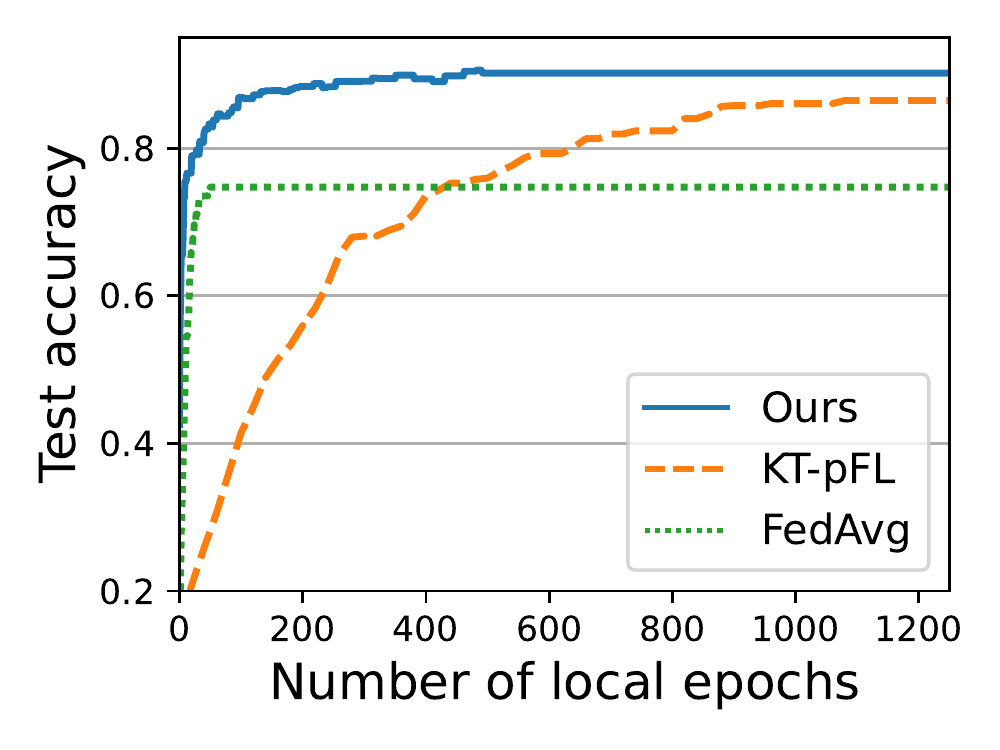}
		\caption{Fashion-MNIST}
	\end{subfigure}
	\begin{subfigure}{.33\textwidth}
		\includegraphics[width=\linewidth]{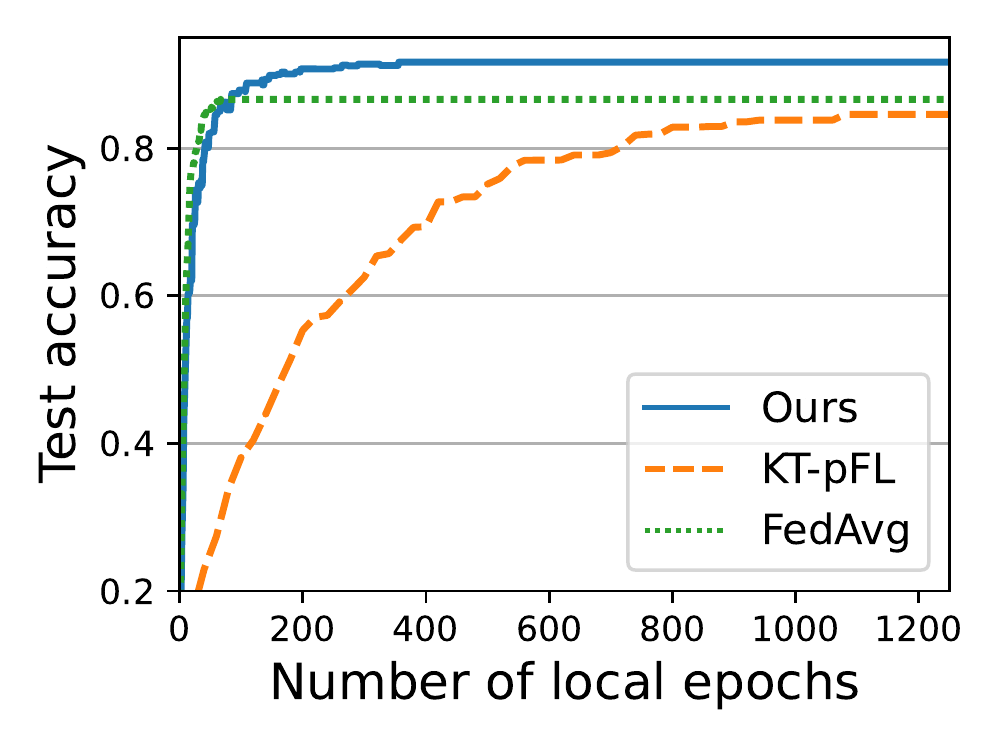}
		\caption{EMNIST}
	\end{subfigure}
	\caption{Average test accuracy of 100 homogeneous models. Data were distributed by Dirichlet distribution (Dir(0.5)) and for each communication round, the client sampling rate was 0.1.}
	\label{fig:homo_100}
\end{figure*}
In this section, we demonstrate the generalization performance of FedClassAvg on heterogeneous models and compare it with a baseline task, FedProto~\cite{FedProto}, and KT-pFL~\cite{ktpfl} algorithm.
Twenty clients with different models were trained on three benchmark datasets.
For baseline, KT-pFL and FedClassAvg we used ResNet-18, ShuffleNetV2, GoogLeNet, and AlexNet with modifications at final fully connected layers.
Meanwhile, FedProto used different model structures since it requires the prototypes to be the same dimensions. 
Hence we used the same model heterogeneity schemes as described in \cite{FedProto}: two-layer convolutional neural networks with different output channels for Fashion-MNIST and EMNIST datasets, and ResNet-18 with different strides for CIFAR-10~\cite{FedProto}.
Please note that FedProto experiments assume less heterogeneous client models than the other methods.

For the baseline task, we measured the average test accuracy when each local model was trained using only the local data.
Personalized federated learning algorithms were evaluated based on average performance on test datasets consistent with local data distributions.
However, to ensure the clients benefit from federated learning, we should determine whether there is a performance improvement rather than training with local updates only.

Figures~\ref{fig:hetero_labeldir} and \ref{fig:hetero_twoclass} shows the learning curves.
Note that the x-axis is the number of local epochs, not the number of communication rounds for fair comparison, because KT-pFL requires 20 epochs of local updates whereas the others were trained for a single epoch.
\cready{
In most cases, FedClassAvg converges faster than KT-pFL with higher test accuracies, except for Figure~\ref{fig:hetero_twoclass_cifar10}.
We believe it is because KT-pFL initializes local models by training one local epoch while FedClassAvg does not. 
When the data becomes complicated, local models trained without initialization might take time from trials and errors. 
Still, the final model performance after convergence is consistently higher in the proposed method.
}

The average and standard deviations of final test accuracies are listed in Table~\ref{tab:hetero_testacc}.
As shown in the table, our proposed method showed consistent superiorities in average test accuracy.
Furthermore, the standard deviations of client accuracies is mostly smaller than other methods, meaning that FedClassAvg guarantees high and consistent accuracies among clients.
\subsection{Homogeneous federated learning}
In this section, we evaluate the FedClassAvg under two scenarios for federated learning on homogeneous models.
In the first scenario, the clients train in the same manner as heterogeneous model training.
And in the second scenario, clients share all the model weights, including the feature extractor parts.
In the case of FedClassAvg, both classifier and feature extractor weights are aggregated after local client updates, and only proximal regularization regularizes the classifier weights.
For KT-pFL on homogeneous models, each client model maintains the personalized global model on the server.
After the local client updates, the model weights are linearly combined with the corresponding global models.
The aggregated global model weights are then used for knowledge transfer instead of soft predictions for public data.

For client models, we used ResNet-18 backbone layers as feature extractors, followed by single-layer fully connected layers as classifiers.
Data heterogeneity was assumed to be Dir(0.5).
We measured the average test accuracy of the clients for the local test datasets when the total number of clients was 20 and 100.
The client sampling rate was set to 1 for 20 clients and 0.1 for 100 clients.
The comparison results between FedAvg, KT-pFL and the proposed method are summarized in Table~\ref{tab:homo_testacc} and Figures~\ref{fig:homo_20} and \ref{fig:homo_100}.
\cready{
FedAvg and FedProx are algorithms defined only when the entire model weights are exchanged (the second scenario). 
Therefore, in the first scenario where only the FC layers are exchanged, the amount of information exchanged with each other is significantly smaller than that of FedAvg or FedProx, so the model performance is inevitably inferior. 
Thus, comparing the second scenario performances with those is fair, and as suggested in the paper, FedClassAvg consistently outperforms the other algorithms. 
Additionally, comparing the first scenario performance of the proposed method and KT-pFL, the proposed method consistently outperforms the competitors.
}
\section{Discussion}
\label{sec:discussion}
\subsection{Ablation study}
We conducted ablation studies to determine the effects of the building blocks in the proposed method, and the results are presented in Table~\ref{tab:ablation}.
The training performance improves by using only the contrastive loss, which stabilizes the decision boundary.
However, contrastive loss can separate feature-space representations too far to learn the collaborative knowledge.
Hence, models should be corrected with proximal regularization as suggested in Table~\ref{tab:ablation}.
In all cases, the performance of the model was best when the proposed classifier averaging, proximal regularization, and contrastive loss were all used.
\begin{table}[!t]\centering
\caption{Ablation study (CA: classifier averaging, PR: proximal regularization, and CL: contrastive loss). }\label{tab:ablation}
\begin{tabular}{lccccc}\toprule
\textbf{Data} &\textbf{CA} &\textbf{+PR} &\textbf{+CL} &\textbf{+PR, CL} \\\midrule
CIFAR-10 &0.615 &0.6311 &0.7509 &\textbf{0.7670} \\
Fashion-MNIST &0.8578 &0.8971 &0.924 &\textbf{0.9303} \\
EMNIST &0.915 &0.8993 &0.9186 &\textbf{0.9305} \\
\bottomrule
\end{tabular}
\end{table}
%
%
\subsection{Feature extraction}
\begin{figure*}[!htp]
\begin{subfigure}{.33\linewidth}
    \centering
    \includegraphics[width=\textwidth]{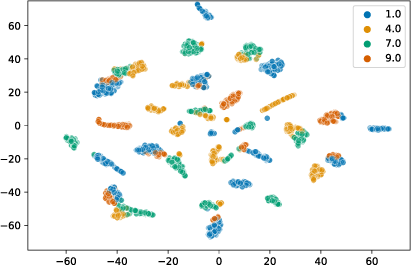}
    \label{fig:tsne_cifar10_noup_labels}
\end{subfigure}
\begin{subfigure}{.33\linewidth}
    \centering
    \includegraphics[width=\textwidth]{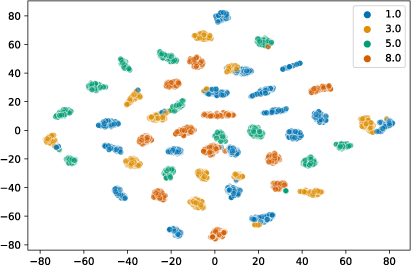}
    \label{fig:tsne_fmnist_noup_labels}
\end{subfigure}
\begin{subfigure}{.33\linewidth}
    \centering
    \includegraphics[width=\textwidth]{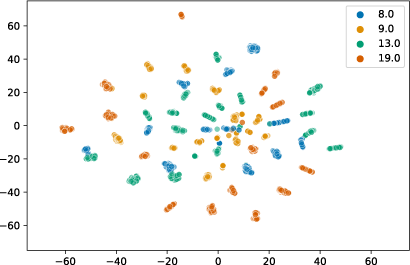}
    \label{fig:tsne_emnist_noup_labels}
\end{subfigure}
\begin{subfigure}{.33\linewidth}
    \centering
    \includegraphics[width=\textwidth]{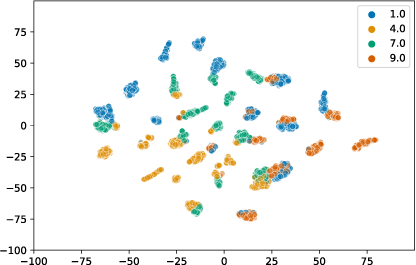}
    \label{fig:tsne_cifar10_labels}
    \caption{CIFAR-10}
\end{subfigure}
\begin{subfigure}{.33\linewidth}
    \centering
    \includegraphics[width=\textwidth]{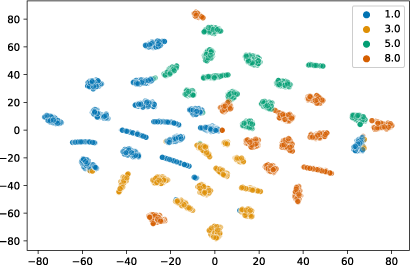}
    \label{fig:tsne_fmnist_labels}
    \caption{Fashion-MNIST}
\end{subfigure}
\begin{subfigure}{.33\linewidth}
    \centering
    \includegraphics[width=\textwidth]{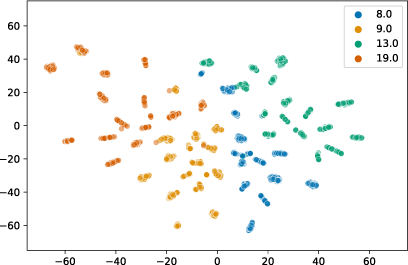}
    \label{fig:tsne_emnist_labels}
    \caption{EMNIST}
\end{subfigure}
\caption{t-SNE results of the trained feature representations: (top) baseline, (bottom) proposed.}
\label{fig:tsne}
\end{figure*}
Figure~\ref{fig:tsne} shows the t-SNE~\cite{tsne} results for the feature representations.
It shows the similarity of the feature representations from each client model after training with FedClassAvg.
The features were extracted from the models trained with the Fashion-MNIST dataset drawn from 1,000 sampled test images.
Top figures in Figure~\ref{fig:tsne} shows the t-SNE of the extracted features of the baseline experiment, when the clients only trained the local dataset without model or knowledge exchange.
Bottom figures in Figure~\ref{fig:tsne} shows the t-SNE results of the extracted features from the client models trained by FedClassAvg.

The top figures in Figure~\ref{fig:tsne} show that the features on the same clients are gathered but rather randomly on labels when trained separately.
However, as shown in botttom figures in Figure~\ref{fig:tsne}, the features with the same label are located close to each other, even for different clients when trained with FedClassAvg.
In particular, it can be seen that the client cluster is split in order to aggregate feature representations of the same label among different clients.

In summary, FedClassAvg gathers the feature-space representations of the same labels.
This makes the heterogeneous clients to share the knowledge learned from local training easily, resulting in better training performance than other methods.
\subsection{Position information of extracted features}
As another way to determine whether the feature extraction is similar among clients, we measured the similarity of the positional attribution of each feature.
Unlike convolutional layers, the input feature position is important for fully connected layers.
If we measure the unit attribution at the classifier layer, the important units should be similar if clients learn similar feature-space representations.
Hence, we use layer conductance~\cite{layer_conductance} to measure the attributes of the classifier. 
Then, we converted the conductance vector to a rank score for comparison with different clients.

\begin{figure}[t]
    \centering
    \includegraphics[width=\linewidth]{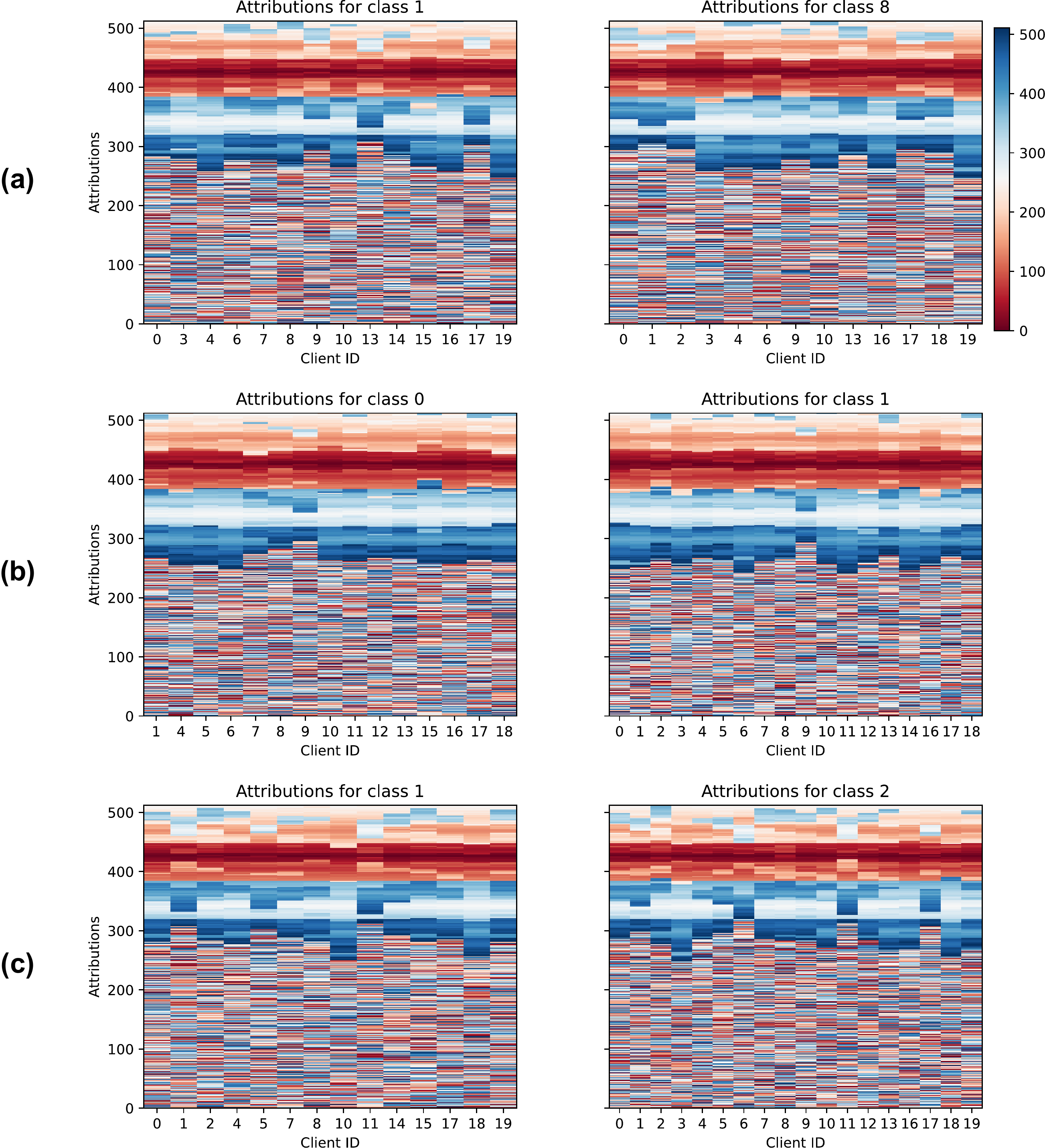}
    \caption{Layer conductance comparison at the fully connected layer of each client model for different datasets. Sampled labels with most correct clients for datasets (a) CIFAR-10, (b) Fashion-MNIST, and (c) EMNIST}
    \label{fig:conductance}
\end{figure}

Figure~\ref{fig:conductance} shows the heatmap result of the attribution rank scores of the eight clients that correctly classified an image.
The x-axis represents the client ID, and the y-axis represents the layer conductance score of each unit.
The backbones of clients 0, 4, 8, 12, 16 are ResNet-18, clients 1, 5, 9, 13, 17 are ShuffleNetV2, clients 2, 6, 10, 14, 18 are GoogLeNet, and clients 3, 7, 11, 15, 19 are AlexNet.
Because we implemented the representation feature dimension as 512, 512 units were placed sequentially on the y-axis.
From Figure~\ref{fig:conductance}, units have a similar attribution rank score in general.
Therefore, despite model heterogeneity, clients trained with FedClassAvg share similar tendency in unit importance in terms of the unit positions.

\subsection{Communication cost and computation complexity}
\begin{table}[!t]\centering
\caption{Communication cost comparison between model sharing (ResNet-18), KT-pFL with public data and classifiers of FedClassAvg in CIFAR-10 training.}\label{tab:comm_cost}
\begin{tabular}{lcccc}\toprule
&ResNet-18 &KT-pFL &\textbf{Proposed} \\\midrule
Comm. cost &43.73 MB &8.9 MB &\textbf{22 KB} \\
\bottomrule
\end{tabular}
\end{table}
\cready{
The communication efficiency of FedClassAvg with the existing methods is described in Table~\ref{tab:comm_cost}.
}
Table~\ref{tab:comm_cost} lists the communication cost required for a client at a single communication round.
Model sharing methods includes the algorithms that require clients to send the entire models to the servers, such as FedAvg and FedZKT.
We measured the communication cost of ResNet-18 and proposed method as the size of saved PyTorch model state\_dict file.
KT-pFL communicates with public data and their soft predictions.
Hence, we estimated the communication cost of KT-pFL as the size of 3,000 instances of public dataset because the size of soft predictions are negligible.
As shown in Table~\ref{tab:comm_cost}, our proposed method is superior in communication efficiency than the other methods.

In addition, FedProto transmits prototypes of 4K units whereas FedClassAvg demands $(512 \times 10)$ units of weight transmission for each client in any dataset with 10 classes.
However, as revealed in Table~\ref{tab:hetero_testacc}, FedClassAvg significantly outperforms the training performance with slight increase in communication.
Therefore, our proposed method is the most communication-efficient with the highest training capabilities among heterogenous clients.

\cready{
The communication cost and its scaling with the growing model complexity are related to the size of classifiers. 
If the number of hidden units in the last FC layer is $N$, the communication cost will grow as $O(N)$.
The computational complexity in local client updates is the same as conventional CNN training. 
In global classifier updates, the model complexity of the global classifier is proportional to the weight dimension of the final FC layer, and not affected by the other model complexity factors. 
Therefore, if the number of hidden units in the last FC layer is $N$, the computational complexity of the global parameter update is $O(N)$.
}

\section{Conclusion}
In this study, we propose a novel personalized federated learning method called FedClassAvg to tackle model heterogeneity.
Through extensive experiments, we demonstrated that FedClassAvg outperformed existing studies.
The primary limitation of this study is that it assumes the client classifiers to have the same architecture, which can affect the model performance a bit.
Combining knowledge transfer techniques or prototype training with our method can bring effective enhancements.
In future work, the proposed method needs to confirm whether it is valid for tasks other than image classification.
In addition, combining other state-of-the-art un/semi-supervised contrastive losses might improve FedClassAvg.

\begin{acks}
\cready{
    This work was supported by SNU-LG Chem DX Center and LG Innotek. The authors would like to thank the anonymous reviewers of this manuscript for their evaluation efforts. 
}
\end{acks}

\bibliographystyle{ACM-Reference-Format}
\bibliography{main}

\newpage
\setcounter{section}{0}
\setcounter{table}{0}
\setcounter{figure}{0}
\renewcommand\thesection{A}
\section{Reproducibility Appendix}
\textit{\textbf{Computational Artifacts}: Yes}

\subsection{Artifact Description Details}
\textit{\textbf{Summarize the experiments reported in the paper and how they were run.}}\\
Our experiments include supervised training of convolutional neural networks across multiple clients implemented using PyTorch 1.9.1 and MPICH 3.2.3. We used a cluster with fifteen nodes, ten with GeForce GTX 1080 Ti GPUs and five with GeForce GTX 2080 Ti GPUs. All nodes have Intel(R) Core(TM) i7-4790 CPU @ 3.60GHz CPUs and 32GB memory. The exact numbers in experimental results may slightly vary from run to run in different environments because of inherent randomness and  nondeterministic implementations in PyTorch. The relative performance of the different algorithms compared should remain the same.

\vspace{1em}

\noindent
\textit{Software Artifact Availability:}\\
Available at https://github.com/hukla/FedClassAvg.

\end{document}